\documentclass[letterpaper, 10 pt, conference]{ieeeconf}  

\IEEEoverridecommandlockouts                              

\usepackage[nocompress]{cite}

\usepackage{amsmath,amssymb,amsfonts}
\usepackage{algorithmic}

\let\labelindent\relax
\usepackage{enumitem}

\newlist{compactitem}{itemize}{1}
\setlist[compactitem]{
    label=\textbullet,
    leftmargin=*,
    topsep=2pt,
    itemsep=1pt,
    parsep=0pt,
    partopsep=0pt
}

\usepackage{graphicx}
\usepackage{textcomp}
\usepackage{xcolor}
\usepackage{booktabs}
\usepackage{multirow}
\usepackage{makecell}
\usepackage{xcolor}
\usepackage[hidelinks]{hyperref}
\usepackage[nameinlink,noabbrev]{cleveref} 

\crefname{figure}{Fig.}{Figs.}
\Crefname{figure}{Fig.}{Figs.}
\crefname{table}{Table}{Tables}
\Crefname{table}{Table}{Tables}
\crefname{section}{Sec.}{Secs.}
\Crefname{section}{Sec.}{Secs.}

\crefname{subsection}{Sec.}{Secs.}
\Crefname{subsection}{Sec.}{Secs.}
\crefname{subsubsection}{Sec.}{Secs.}
\Crefname{subsubsection}{Sec.}{Secs.}

\def\BibTeX{{\rm B\kern-.05em{\sc i\kern-.025em b}\kern-.08em
    T\kern-.1667em\lower.7ex\hbox{E}\kern-.125emX}}

\title{\LARGE \bf
AHEAD: Anticipatory Hand-Driven Teleoperation\\ via Human Intent Prediction
}

\author{Seok Joon Kim$^{1}$, Joonho Lee$^{2}$, Federica Spinola$^{3}$, Taein Kwon$^{4}$, and Mohsen Moghaddam$^{5}$%
\thanks{$^{1}$Georgia Institute of Technology. \textit{seokjoonkim@gatech.edu}}%
\thanks{$^{2}$Neuromeka Ltd. \textit{joonho.lee@neuromeka.com}}%
\thanks{$^{3}$INRIA. \textit{federica.spinola@inria.fr}}%
\thanks{$^{4}$University of Oxford, UK. \textit{taein@robots.ox.ac.uk}}%
\thanks{$^{5}$Georgia Institute of Technology. \textit{mohsen.moghaddam@gatech.edu}}%
}

\begin{document}

\maketitle
\thispagestyle{empty}
\pagestyle{empty}

\begin{abstract}
Direct hand-driven teleoperation maps an operator’s hand motion to robot end-effector commands at every frame, enabling precise control, but it requires constant monitoring and correction during approach, grasp, and placement, which can be slow and fatiguing. For repetitive pick-and-place tasks, supervisory (goal-based) teleoperation simplifies this process: the operator specifies goals/waypoints, and the robot executes the motion using planning algorithms. Yet, this introduces latency, as the robot must wait for the next command before it can plan and act. \textit{How can we reduce robot reaction time while lowering operator workload?} To tackle this question, we present \textbf{AHEAD}, a real-time VR teleoperation system that anticipates operator intent to enable proactive, hand-driven control. In a digital twin, the operator performs pick-and-place naturally, using hand motion to convey high-level commands rather than a continuous robot trajectory. AHEAD processes a short window of 3D hand and head signals together with scene context through an attention-based classifier to predict the intended grasp object and placement slot. A state machine converts intent predictions into stable robot goals, enabling early motion while remaining stable under noisy predictions and corrective hand movements. AHEAD's intent prediction module achieves Top1 accuracy: 76\% for grasp objects and 76\% for target slots. Moreover, our user study shows AHEAD reduces robot reaction latency by 0.6 s (object) and 1.4 s (slot) relative to baselines. Participants also reported lower operator load, indicating faster robot responses while maintaining low operator effort in practice.
\end{abstract}


\section{Introduction}
\label{sec:intro}

Teleoperation allows robots to operate under human supervision when full autonomy is not yet reliable or available. Yet, current teleoperation systems still face a fundamental trade-off. The literature \cite{nationalresearch, SpaceTeleop} commonly distinguishes \emph{direct} control, which demands continuous operator attention and repeated correction, from \emph{supervisory} (goal-based) control, which reduces operator burden but forces the robot to wait for the next explicit command. This trade-off is especially relevant for pick-and-place, where continuous manual teleoperation can be unnecessarily demanding, while passive goal-based interfaces can introduce lag at each subgoal. This raises the question of whether robots can begin acting earlier by predicting operator intent from ongoing motion.

More specifically, direct teleoperation maps operator input to robot commands at high rates. In hand-tracked settings, this often means retargeting hand poses at every frame under joint-limit, morphological gaps and collision constraints~\cite{anyteleop, opentelevision, openteach}. This paradigm supports fine-grained control, for example in surgical teleoperation\cite{davinci}, and data collection for downstream learning, where per-timestep action information matters. However, this paradigm requires operators to monitor execution and repeatedly correct drift, overshoot, and misalignment. This increases both cognitive and physical workload and can yield slow or jerky motion. For pick-and-place tasks, we argue that this level of continuous control is often unnecessary. Instead, goal-based teleoperation uses task-level inputs, such as selecting a grasp object and a placement target, letting the robot plan and execute the motion. This approach has been shown to reduce operator burden compared with direct control\cite{teleop_hard2}. However, standard supervisory interfaces typically begin planning only after the operator explicitly specifies the next waypoint via a UI, ray-cast, or confirmation action\cite{multimodalui, explicit}. This adds non-negligible idle time to the robot's response and makes interaction feel slow and laggy. We refer to this delay as the robot's \textit{reaction time}, defined here as the time between the onset of the operator's reach and the onset of robot motion.

\begin{figure}[t!]
    \vspace{8pt}
    \centering
    \includegraphics[width=\columnwidth]{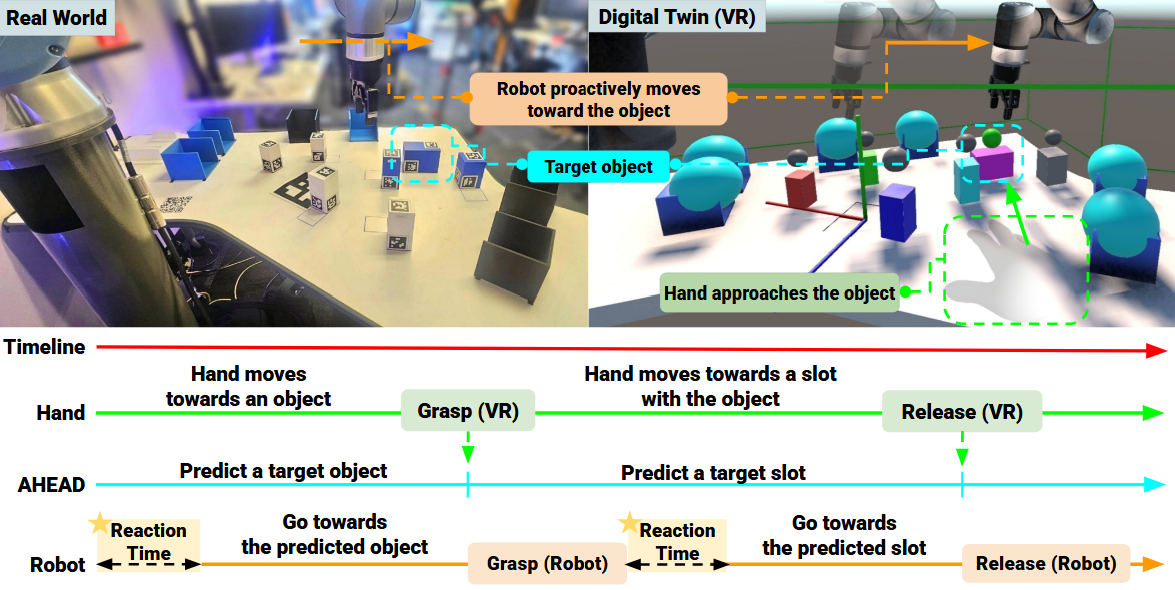}
    \caption{\textbf{Top:} Real-world workspace and VR digital twin used in AHEAD. \textbf{Bottom:} The timeline highlights AHEAD’s key goal: reducing robot reaction time. As an operator’s hand reaches toward an object and then toward a slot in VR, AHEAD predicts the intended object and target slot from recent hand motion, and initiates robot motion toward the predicted goal early without explicit selection. After the VR grasp and release signals, the robot executes grasp and place at the selected object and slot.}
    \label{fig:main_front_teaser}

\end{figure}

We address this limitation with \textit{anticipatory goal-based teleoperation}. In this paradigm, the operator provides goal-level input through natural hand motion, so the interaction remains goal-based and low-burden. Intent prediction \cite{gaze-goal,landmark_based, penco2024mixedreality} makes this possible in teleoperation because the operator's ongoing behavior (hand motion, gaze, body orientation) often reveals the next subgoal without explicitly issuing a command. The robot can therefore start planning and moving as soon as the operator's intent becomes sufficiently reliable. This preserves the structure of goal-based teleoperation while recovering some of the responsiveness usually associated with direct control.

To realize this idea, we propose \textbf{AHEAD}, a virtual reality (VR) teleoperation system for anticipatory goal-based teleoperation based on human intent prediction. AHEAD provides an immersive digital twin of the remote workspace, allowing the operator to control robots from remote locations. In VR, the operator uses natural hand motions, including reaching, grasping and releasing, to implicitly express high-level goals, such as what to pick and where to place it. Unlike 2D image-based intent prediction pipelines \cite{EgoPat3Dv2, EgocentricActionRecognition, HandsOnVLM}, AHEAD predicts intent directly from wearable 3D signals that modern VR and augmented reality (AR) devices already provide: 3D hand joints and head pose \cite{MetaHandTrackingOverview, MRTK3HandTracking}. We combine these signals with an instance-agnostic scene model, representing graspable objects as cuboids and placement targets as spheres. We predict intent through phase-specific attention-based classifiers: the grasp object during reach, then the placement target after grasp. AHEAD then uses a distance-gated policy through a state machine to convert these predictions into action: it can issue a conservative early preview when intent is still uncertain, and commit to the target once intent becomes reliable. By starting robot motion without explicit operator commands, AHEAD reduces reaction time while preserving the low cognitive burden of goal-based teleoperation. Our contributions are three-fold: 

\begin{enumerate}
\item \textbf{Anticipatory goal-based teleoperation for immersive pick-and-place:}
We present \textbf{AHEAD}, a real-time VR digital twin pick-and-place teleoperation system with a reactive yet stable robot controller based on distance-gated preview-and-commit motion.

\item \textbf{Hand-driven intent prediction:}
We propose an attention-based intent predictor that infers grasp objects and placement slots from streaming 3D hand joints, head pose, and scene context, achieving Top$1$/Top$3$ accuracy of 76/90\%  for object and 76/91\% for slot.

\item \textbf{Human-subject validation of reduced reaction time and workload:}
From a 15-participant study, we show AHEAD reduces robot reaction time by 0.6\,s (object) and 1.4\,s (slot) relative to the baselines, while improving user ratings and lowering NASA-TLX \cite{NASATLX} workload.
\end{enumerate}

\section{Related Work}
\label{sec:related_work}
\noindent\textbf{A. Robot Teleoperation:}
Teleoperation spans a spectrum from \emph{direct} control, which retargets tracked human motion to robot commands at high rate, to \emph{supervisory/goal-based} control, where an operator specifies step-wise goals and a robot executes motion primitives. Direct teleoperation enables immediate, embodied control and scalable demonstration collection utilizing VR \cite{openteach,opentelevision} and hands \cite{anyteleop}, but full-rate retargeting can be sensitive to network latency, kinematic mismatch, and operator micro-corrections, amplifying jitter and cognitive workload \cite{teleop_hard2, penco2024mixedreality}. Hardware-mediated interfaces such as GELLO \cite{GELLO} and HOMIE \cite{ben2024homie} improve proprioceptive coupling and demonstration quality, but require additional hardware to interact with an environment. Some systems improve usability via \emph{preview-before-execute} teleoperation (e.g., overlaying a spatially aligned virtual robot before committing actions)~\cite{telepreview}. In contrast, goal-based teleoperation reduces operator burden by shifting interaction to UI-mediated procedures or task-level commands with automated execution~\cite{tasklevelauthoring,explicit}. However, these interfaces often remain passive: planning and execution typically begin only after the operator explicitly specifies (and often confirms) the next goal or plan~\cite{tasklevelauthoring}. Our approach follows a goal-based teleoperation paradigm while moving toward continuous, direct interaction: operators use natural hand motion, and the system proactively reacts within a goal-based framework without requiring explicit goal specification in VR.

\noindent\textbf{B. Intent Prediction from hands:} A large body of work predicts goals and future interactions from hand motion and scene context. In egocentric vision (2D images), methods predict 3D action targets \cite{EgoPat3Dv2}, model hand--object contact/state for action understanding \cite{EgocentricActionRecognition, holoassist}, forecast future hand motion and interaction regions \cite{liu2022joint}, and predict 3D hand trajectories even when hands are not visible \cite{invisible_ego_hand}. Vision-language models have also been applied to hand--object interaction prediction \cite{HandsOnVLM}. EgoPAT3Dv2 is closest to our goal of anticipatory robot assistance in that it predicts 3D action targets for downstream robot execution from egocentric images in an \emph{in-situ} AR setting. In contrast, AHEAD targets \emph{remote} teleoperation through a VR digital twin and uses tracked \emph{3D hand joints} as the primary real-time intent signal. Complementary VR work predicts future \emph{3D hand trajectories} directly from tracked hand kinematics \cite{sopredictable} for early anticipation. In contrast to vision-centric approaches, we combine tracked \emph{3D hand} and head motion with a geometric encoding of scene entities to enable low-latency intent prediction for proactive goal-based teleoperation.

\section{Method}
\label{sec:method}
This section presents \textbf{AHEAD}, a VR teleoperation system deployed in a digital twin workspace. It predicts operator intent to proactively initiate robot motion and reduce robot \emph{reaction time}, shown as star markers in \cref{fig:main_front_teaser}.

\subsection{Problem Setup and Representation}
\label{sec:problem_setup}

We consider a pick-and-place teleoperation task in which the robot operates in a physical workspace while the operator interacts remotely through a mirrored VR digital twin (see top row of \cref{fig:main_front_teaser}). Using a Meta Quest 3, the operator views the digital twin and controls the robot through natural reach-and-place motions. We consider single-handed tasks and model right-hand motion. However, the same pipeline can support left-handed use by mirroring the right-hand data or by collecting additional left-hand data.

To maintain real-to-sim consistency, we calibrate the robot base pose to a fixed world frame using a ChArUco \cite{ArUco} board and estimate object poses in real time from ArUco \cite{ArUco} markers attached to each object (six markers, one per face). We use markers for object pose estimation to ensure stable pose tracking, and focus on intent prediction and interaction time instead. ArUco tracking can be replaced by state-of-the-art markerless pose estimators when needed\cite{Pos3R}. At each time step $t$, the system receives a short temporal window of hand and head motion together with scene entities: (i) $N$ graspable objects represented as cuboids ($N \le 10$) and (ii) $M$ target locations (slots) represented as spheres ($M \le 10$). We set $N,M \le 10$ to avoid visual clutter in the VR scene.

\subsubsection{Hand and Head Inputs}
\label{subsec:hand_and_head_inputs}

We encode the right hand at time $t$ as $\mathbf{h}_t \in \mathbb{R}^{63}$ using the 3D positions of 21 joints. From the 26-joint OpenXR hand model~\cite{OpenXRHand}, we drop metacarpal joints and use a 21-joint set: wrist, palm, three thumb joints, and four joints for each remaining finger (index--pinky). We encode the headset state as $\mathbf{g}_t=[\mathbf{p}^{\text{head}}_t;\mathbf{d}^{\text{head}}_t]\in\mathbb{R}^{6}$, where $\mathbf{p}^{\text{head}}_t\in\mathbb{R}^{3}$ is head position and $\mathbf{d}^{\text{head}}_t\in\mathbb{R}^{3}$ is a unit head-forward ray vector. Since Meta Quest 3 does not support eye-gaze tracking, we use $\mathbf{d}^{\text{head}}_t$ as a proxy for gaze direction. The intent predictor operates on fixed-length sequences of the most recent $T$ frames. We define the hand and headset input sequences at time $t$ as:
\[
\begin{aligned}
\mathbf{H}_{t} &= [\mathbf{h}_{t-T+1}, \ldots, \mathbf{h}_{t}] \in \mathbb{R}^{T \times 21 \times 3}, \\
\mathbf{G}_{t} &= [\mathbf{g}_{t-T+1}, \ldots, \mathbf{g}_{t}] \in \mathbb{R}^{T \times 6}.
\end{aligned}
\]

We use a 0.5 s temporal horizon with $T=30$ samples at the nominal 60 Hz Meta Quest 3 hand/head tracking rate and run inference at every frame $t$ using a sliding window. Because tracking FPS can vary, we maintain a rolling 0.5 s buffer of hand/head samples and resample it to exactly $T$ frames using uniform time sampling with interpolation when needed. We linearly interpolate translational quantities (hand joints and head positions) and apply SLERP~\cite{SLERP} to the head-forward ray direction.

\subsubsection{Scene entity inputs}
\label{subsec:scene_entity_inputs}

We represent each object as a cuboid descriptor $\mathbf{o}_i$ consisting of its centroid and eight corners:
$\mathbf{o}_i = [\mathbf{c}_i, \mathbf{p}_{i,1}, \ldots, \mathbf{p}_{i,8}] \in \mathbb{R}^{27}$,
where $\mathbf{c}_i \in \mathbb{R}^3$ is the cuboid centroid and
$\mathbf{p}_{i,k} \in \mathbb{R}^3$, $k=1,\ldots,8$, are its corners.
We represent each slot as a sphere descriptor
$\mathbf{s}_j = [\mathbf{u}_j, r_j] \in \mathbb{R}^4$,
where $\mathbf{u}_j \in \mathbb{R}^3$ is the sphere center and
$r_j \in \mathbb{R}$ is its radius. We use cuboids to keep objects instance-agnostic: any object can be reduced to a 3D bounding box, avoiding usage of mesh/point cloud encoders that require more data and computation. We represent slots as spheres so the operator can approach slots from any direction, without requiring a specific approach direction.

\subsubsection{Head-centric coordinate frame}
\label{subsec:head_centric}

We express all hand joints and scene entities in a head-centric frame to reduce dependence on global coordinates. World-frame positions depend on the workspace origin and object layout, which can encourage the model to overfit to the data collection setup. A head-centric representation instead emphasizes \emph{relative} head-to-hand and hand-to-entity geometry, making the model more \emph{dataset-setup agnostic} and improving transfer across environments. For each sliding window ending at time $t$, we anchor the frame to the first head pose $(\mathbf{R}_{\text{head},t_0}, \mathbf{t}_{\text{head},t_0})$ with $t_0=t-T+1$. Any world-frame point $\mathbf{x}_\tau$ in the window is transformed as $\tilde{\mathbf{x}}_\tau=\mathbf{R}_{\text{head},t_0}^{\top}(\mathbf{x}_\tau-\mathbf{t}_{\text{head},t_0})$, $\tau\in\{t_0,\ldots,t\}$. For directional features, we apply rotation only. The head-ray direction becomes $\tilde{\mathbf{d}}^{\text{head}}_\tau=\mathbf{R}_{\text{head},t_0}^{\top}\mathbf{d}^{\text{head}}_\tau$.

\subsubsection{Intent outputs}
\label{subsec:intent_output}

The model predicts two categorical intents: which object to pick next and which slot to place it in. We switch from object to slot prediction when a grasp signal is received from the Meta \textit{TouchHandGrab} callback~\cite{MetaTouchHandGrab}. Before this callback signal, the model outputs pick probabilities over the N objects. After the grasp signal, it outputs place probabilities over the M slots.

\subsection{Dataset Collection}
\label{sec:dataset_collection}

We collect a dataset of pick-and-place demonstrations in VR using a Meta Quest~3. The participants are 3 males/2 females, all right-handed and familiar with VR. Each trajectory is a complete episode spanning \textit{start $\rightarrow$ approach $\rightarrow$ grasp $\rightarrow$ place $\rightarrow$ release} and includes time-synchronized 3D right-hand joint kinematics $\mathbf{h}_t$, head pose $\mathbf{g}_t$, scene entity descriptors $\{\mathbf{o}_i\}_{i=1}^N$ and $\{\mathbf{s}_j\}_{j=1}^M$, and event annotations (\textit{grasp}, \textit{slot-entry}, \textit{release}), recorded at 60~Hz. We use Meta Quest \textit{TouchHandGrab}~\cite{MetaTouchHandGrab} to enable natural grasp and release in VR with hand motions that closely resemble real-world manipulation, and to receive callback events for both \textit{grasp} and \textit{release}.

\subsubsection{Scene generation}
\label{subsec:scene_generation}
\begin{figure}[t]
    \vspace{6pt}
    \centering
    \includegraphics[width=\columnwidth]{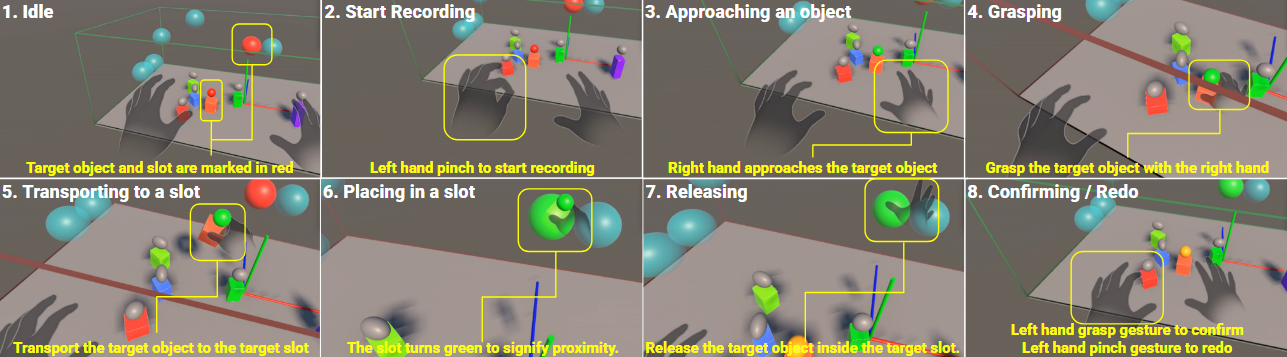}
    \caption{Data collection procedure: A left-hand pinch starts recording (2), then the right hand reaches (3) and grasps (4) the object, and transports (5) and places it in the slot (6). After release (7), operators can redo (left-hand pinch) or advance (left-hand grasp) (8). Yellow boxes highlight key events.}
    \label{fig:dataset_collection_scheme}
\end{figure}

Each scene contains a variable number of entities. We sample the number of objects $N$ and slots $M$ uniformly from 5$ \sim$10. To control the spatial extent of scene layout, we sample positions within a workspace region $\mathcal{B}_{\text{work}}=[x_{\min},x_{\max}]\times[z_{\min},z_{\max}]$ with $x\in[-0.75,0.75]$\,m and $z\in[-0.375,0.375]$\,m. The coordinate frame is shown in \cref{fig:dataset_collection_scheme} (Unity left-hand convention), where $x$ is red, $y$ is green, and $z$ is blue. Participants face the $+z$ direction, and $\mathcal{B}_{\text{work}}$ corresponds to the tabletop region in \cref{fig:dataset_collection_scheme}. Objects are cuboids placed on the tabletop ($y=0$) with randomized yaw and non-overlap constraints. Their centers are sampled from a tighter region $\mathcal{B}_{\text{obj}}$ with $x\in[-0.40,0.40]$\,m and $z\in[-0.35,0.30]$\,m to prevent object--slot collisions during scene generation.
Object dimensions are randomized per instance, with width and length sampled from $[0.040,\,0.065]$\,m and height sampled from $[0.05,\,0.13]$\,m. Slot centers are sampled along the edges of the $x-z$ workspace at varying heights $+y$. We do not use the near edge ($z=z_{\min}$) for the slot generation to avoid occluding objects in the participants' view. We then select $M$ non-overlapping slots per scene. Slot radii are sampled from $[0.10,0.20]$m during data collection and fixed to $0.15$m in the user study (\cref{sec:eval_userstudy}) to match the real slot size shown in \cref{fig:main_front_teaser} and \cref{fig:robot_movement}. 

\subsubsection{Dataset Collection Procedure}
\label{subsec:interactive_trial_protocol}

The intent prediction dataset collection procedure is illustrated in \cref{fig:dataset_collection_scheme}. Each participant completes 40 scenes with 5 trials per scene lasting on average 3.7\,s each, yielding 200 trajectories per participant (1000 sequences total). In each trial, one target object and one target slot are uniformly randomly sampled without replacement from $N$ objects and $M$ slots. Targets are cued in red: the target object by a small sphere above it and the target slot by a red highlight. Non-target objects use gray spheres and non-target slots are shown in cyan. Placing the small sphere above each object keeps the target cue visible even under occlusion. Each trial follows eight steps: after idle, a left-hand pinch starts recording and turns the target object's sphere green (Step~2) to signal recording status. The participant then uses the right hand to approach, grasp, transport, place, and release the object (Steps~3--7). The slot turns green once the object enters a slot (Step~6), indicating that it can be released. The \textit{release} event marks the end of the data sequence, and the system enters an orange confirmation state (Step~8), where a left-hand pinch repeats the trial and a left-hand grasp advances to the next one.

\subsubsection{Data types}
\label{subsec:dataset_types}
Our data collection yields continuous sequences that span the full pick-and-place motion. When we form training examples using a sliding window $T$, each window can fall into one of three phases depending on whether the \textit{grasp} event occurs within the window. We categorize windows that occur entirely before grasp as \textbf{PreGrasp}, windows that contain the grasp event as \textbf{CrossOver}, and windows that occur entirely after grasp as \textbf{PostGrasp}. This distinction is important because we use these subsets to train different prediction heads as explained in \cref{sec:method_model}.

\subsection{Intent Prediction Model}
\label{sec:method_model}

\begin{figure}[b!]
    \centering
    \includegraphics[width=0.9\columnwidth]{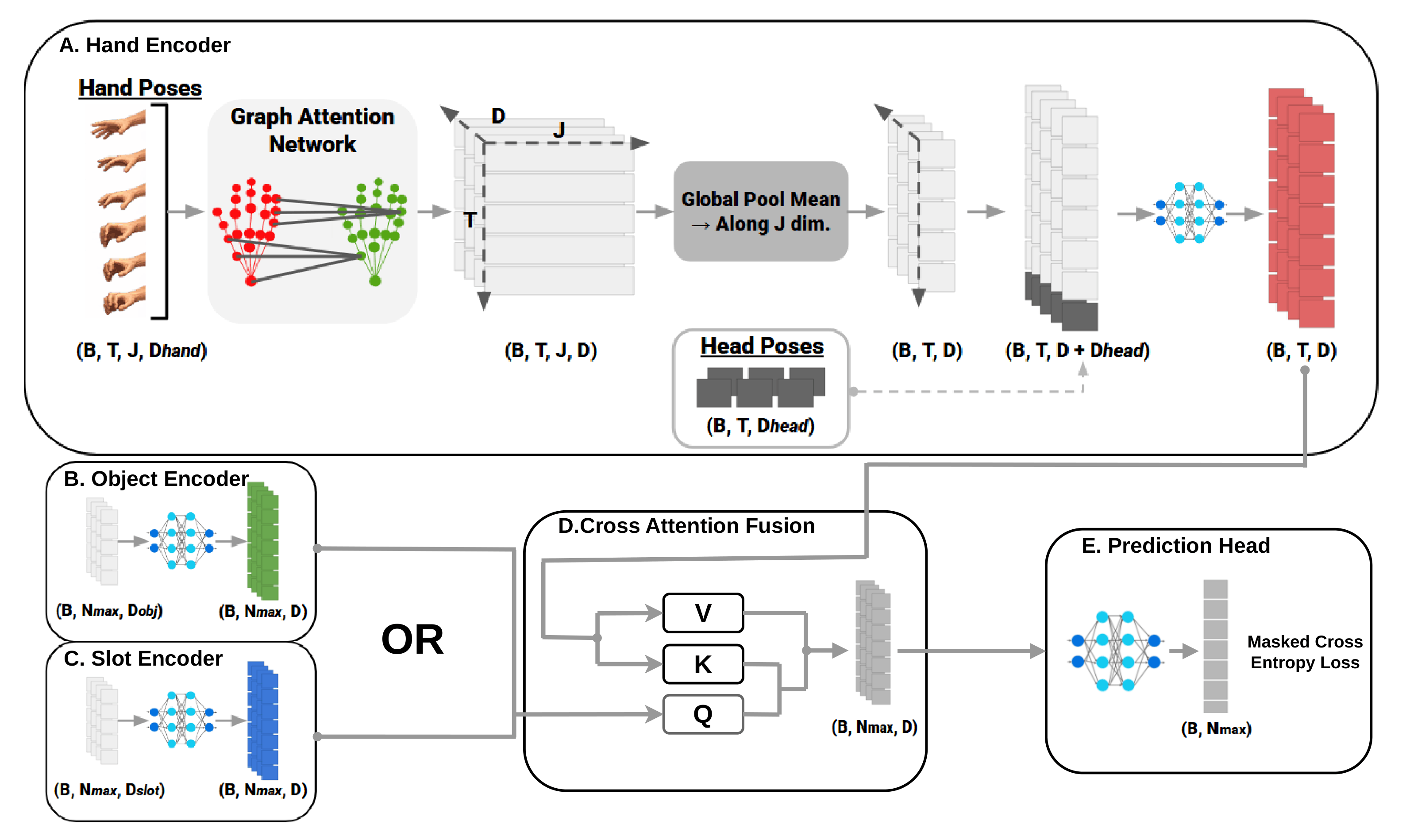}
    \caption{Phase-specific intent prediction model architecture. Depending on the phase, we fuse hand-head features (A) with either object (B) or slot (C) embeddings via cross-attention (D) and predict the target (E).}
    \label{fig:neuralnetwork}
\end{figure}

AHEAD predicts intent from a window of head-centric hand $\tilde{\mathbf{H}_{t}}$ and head $\tilde{\mathbf{G}_{t}}$ motions, and the head-centric entity set $\{\tilde{\mathbf{o}_i}\}_{i=1}^N$ and $\{\tilde{\mathbf{s}_i}\}_{i=1}^M$ ( $\tilde{.}$ represents head-centric coordinates). We formulate hand-driven intent inference as a fixed-size classification problem over up to $N_{\max}=10$ candidate objects or slots. For scenes with fewer candidates, we zero-pad the entity set to this maximum and maintain a binary validity mask, which we apply in attention and in the logits/loss so the padded entries do not affect training. Each phase-specific model then outputs a categorical distribution over valid candidates and is trained with masked cross-entropy. Throughout this section, we will continuously refer to the overview of our model architecture shown in \cref{fig:neuralnetwork}.

\subsubsection{Phase-specific modeling}
\label{subsec:phase-specific-models}
We use two phase-specific intent models that share the architecture in \cref{fig:neuralnetwork}: a hand encoder (A), an entity encoder (B or C), an attention-based fusion module (D), and a prediction head (E). The pre-grasp model uses the object encoder (B) and trains on \textbf{PreGrasp} windows defined in \cref{subsec:dataset_types}, while the post-grasp model uses the slot encoder (C) and trains on \textbf{CrossOver} and \textbf{PostGrasp} windows. At runtime, the system switches from the pre-grasp object predictor to the post-grasp slot predictor when a grasp is detected via the Meta \textit{TouchHandGrab} callback~\cite{MetaTouchHandGrab}. We justify this two-stage design in \cref{sec:eval_offline}.

\subsubsection{Architecture}
\label{subsec:arch}
Throughout this subsection, we omit the batch dimension for clarity. In practice we train the networks over batches of $B$ windows ($B$=100).

\noindent\textbf{Hand encoder (A):}
Given a right-hand window $\tilde{\mathbf{H}}_t \in \mathbb{R}^{T \times J \times D_{hand}}$, we represent each frame $\tau\in\{t_{0},\ldots,t\}$ as a 21-node graph with edges, following the OpenXR hand kinematic tree~\cite{OpenXRHand}. For each frame, a two-layer GATv2 ~\cite{GATv2} produces per-joint features $\mathbf{z}_{\tau} \in \mathbb{R}^{J \times D}$. We average over joints to obtain a per-frame embedding $\mathbf{e}_{\tau} \in \mathbb{R}^{D}$, and stack over time to form $\mathbf{E}_t \in \mathbb{R}^{T \times D}$. We incorporate viewpoint context by concatenating $\mathbf{E}_t$ with the synchronized headset tensor $\tilde{\mathbf{G}}_t \in \mathbb{R}^{T \times D_{head}}$ and passing $[\mathbf{E}_{t};\tilde{\mathbf{G}}_{t}]$ through a Multi-Layer Perceptron (MLP), yielding head-aware hand features $\tilde{\mathbf{E}}_t \in \mathbb{R}^{T \times D}$.

\noindent\textbf{Scene entity encoder (B and C):}
We embed scene objects and slots by stacking the initial object features $\{\tilde{\mathbf{o}_i}\}_{i=1}^N$ and slot features $\{\tilde{\mathbf{s}_i}\}_{i=1}^M$ into $N \times D_{\mathrm{obj}}$ and $M \times D_{\mathrm{slot}}$ input tensors, respectively. We pad them to $N_{\max}$ candidates as needed, and pass the resulting tensors through a 3-layer MLP to obtain object or slot tokens in $\mathbb{R}^{N_{\max} \times D}$.

\noindent\textbf{Attention-based fusion and prediction (D and E):}
We fuse hand-head encodings $\tilde{\mathbf{E}}_t \in \mathbb{R}^{T \times D}$ with scene entity encodings in $\mathbb{R}^{N_{max} \times D}$ using a multi-layer attention block~\cite{AttentionIsAllYouNeed}. The fusion module applies (i) self-attention over the hand sequence (with sinusoidal positional encoding), (ii) self-attention over entity tokens (with sinusoidal positional encoding), and (iii) cross-attention where entity tokens (Q) query the hand sequence (K, V), producing motion-conditioned entity embeddings in $\mathbb{R}^{N_{max} \times D}$. A 2-layer MLP head maps each updated entity embedding to a scalar logit in $\mathbb{R}^{N_{max}}$. We then apply a (masked) softmax to these logits to obtain a categorical distribution over the $N_{\max}$ candidate entities, with invalid padded entries excluded.

\subsubsection{Implementation Details}
Unless otherwise stated, all MLPs and attention blocks use ReLU\cite{relu}, dropout \(p{=}0.1\), and feature dimension \(D{=}128\). MLPs use hidden size \(128\) and attention uses 4 heads. The inputs are \(J{=}21\) hand joints with \(D_{\text{hand}}{=}3\), head features with \(D_{\text{head}}{=}6\), object features with \(D_{\text{object}}{=}27\) and slot features with \(D_{\text{slot}}{=}4\).
For deployment, we export models to ONNX\cite{onnxruntime} and run ONNX Runtime on an NVIDIA RTX PRO 5000 GPU. ORT inference takes $2.60\pm0.32$ms (object) and $2.55\pm0.27$ms (slot) per forward pass, which corresponds to 340--390Hz.

\subsection{Distance-Gated Robot Controller Policy}
\label{sec:method_policy}

\begin{figure}[b!]
    \centering
    \includegraphics[width=0.75\columnwidth]{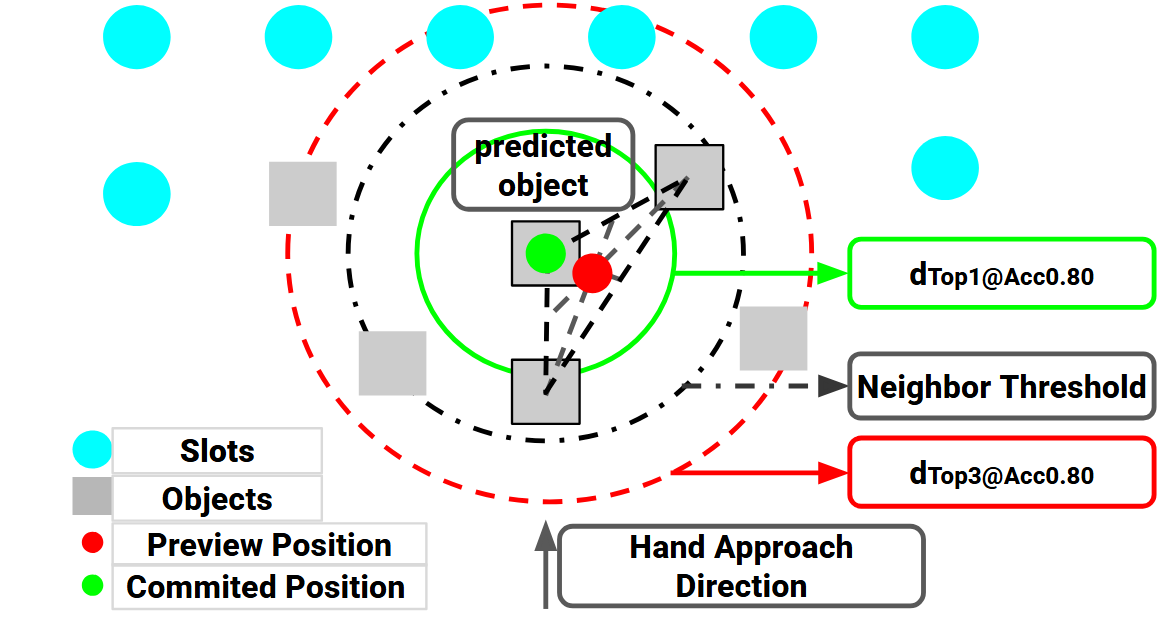}
    \caption{Distance-adaptive proactive robot motion: When the hand enters the neighbor-reliable region $d^{Top3}_{acc=0.8}$, the robot previews above the current neighborhood hypothesis (red dot). When it enters the identity-reliable region $d^{Top1}_{acc=0.8}$, the robot refines to the Top1 object (green dot) and commits, fixating on that object for stability.}
    \label{fig:robotcontroller}
\end{figure}

AHEAD must convert per-frame object or slot predictions into stable, real-time robot commands. Acting directly on the raw predicted target at each frame would make the robot respond quickly, but also overreact to prediction jitter. To balance reactiveness and stability, AHEAD uses a distance-gated controller that commits once the prediction stabilizes. Below, we describe the \emph{pre-grasp} object phase. The same policy applies \emph{post-grasp} for slots.

\subsubsection{Empirical distance thresholds}
\label{subsec:dist_thresh}
Prediction reliability depends strongly on hand--target distance. Therefore, we define two empirical thresholds used in \cref{subsec:policy_logic}: a \emph{neighbor-reliable} distance $d^{\mathrm{Top3}}_{\mathrm{acc}=0.8}$ (red dotted line in \cref{fig:robotcontroller}) and an \emph{identity-reliable} distance $d^{\mathrm{Top1}}_{\mathrm{acc}=0.8}$ (green dotted line in \cref{fig:robotcontroller}). To estimate them, we bin validation segments by palm--target distance and measure (i) Top1 accuracy and (ii) neighbor-aware Top3 accuracy, which counts a prediction as correct if it lies among the three nearest neighbors of the ground-truth target and within radius $r_{\mathrm{nbr}}$ (dotted black line in \cref{fig:robotcontroller}). We then set $d^{\mathrm{Top1}}_{\mathrm{acc}=0.8}$ and $d^{\mathrm{Top3}}_{\mathrm{acc}=0.8}$ to the distances at which the corresponding accuracy exceeds $0.80$ (values reported in \cref{tab:policy_compare}). The accuracy of 0.80 is a design choice that balances robot reaction time with prediction stability. We note that as the hand approaches, predictions converge on a small, consistent set of objects, so the preview–commit logic prevents robot fluctuations by first moving toward the nearby set and then committing above a single object. Also, we use hysteresis: the predicted target is latched when its confidence exceeds $0.65$ and switches only if an alternative candidate's confidence exceeds $0.65$ for $0.1$s (10 frames at 100Hz), avoiding jittery back-and-forth robot motion.

\subsubsection{Distance-gated policy logic}
\label{subsec:policy_logic}
At time $t$, the pre-grasp model outputs a categorical distribution $p_{\mathrm{pre\text{-}grasp}}(\cdot;t)$ over objects, and we take the Top1 hypothesis $i_1(t)=\arg\max_i p_{\mathrm{pre\text{-}grasp}}(i;t)$. Let $\mathbf{p}_{\mathrm{palm}}(t)\in\mathbb{R}^3$ be the palm position, $\mathbf{c}_{i_1(t)}\in\mathbb{R}^3$ the centroid of $i_1(t)$, and $d(t)=\lVert \mathbf{p}_{\mathrm{palm}}(t)-\mathbf{c}_{i_1(t)} \rVert_2$. If $d(t)>d^{\mathrm{Top3}}_{\mathrm{acc}=0.8}$, the robot stays idle. Otherwise, it enters preview mode. We form a gated local ambiguity set
$\mathcal{N}(i_1(t))=\mathrm{Top3NN}(i_1(t))\cap\{k:\lVert \mathbf{c}_k-\mathbf{c}_{i_1(t)} \rVert_2\le r_{\mathrm{nbr}}\}$
with $r_{\mathrm{nbr}}=0.3\,\mathrm{m}$. $\mathrm{Top3NN}(i)$ returns the three nearest objects (including $i$) by centroid distance. The robot previews by moving to a waypoint above the centroid of $\mathcal{N}(i_1(t))$ (red dot in \cref{fig:robotcontroller}), which keeps motion within the local candidate region and enables a reactive commit to an object once the hand reaches the identity-reliable region (green dot in \cref{fig:robotcontroller}).

\subsubsection{Motion primitives}
\label{subsec:motion_primitives}
\begin{figure}[t!]
    \vspace{6pt}
    \centering
    \includegraphics[width=1.0\columnwidth]{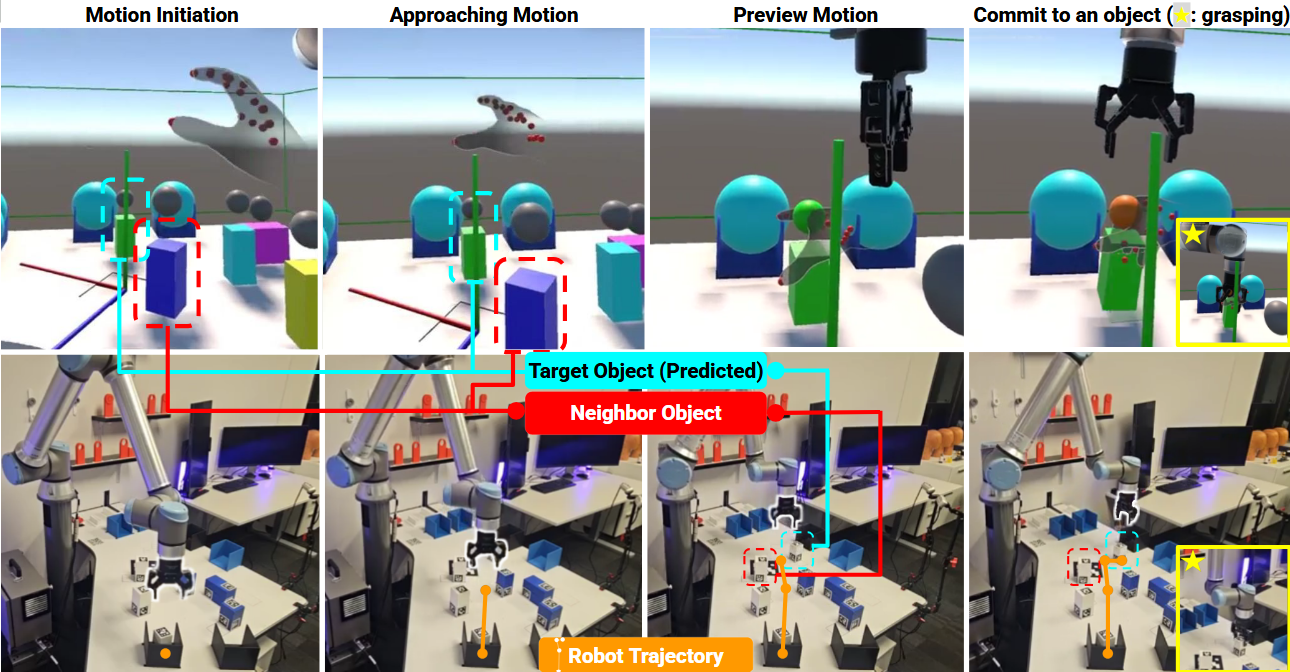}
    \caption{ Hand and robot motion for pre-grasp. As the operator’s hand approaches, AHEAD continuously predicts the target object (cyan outline) among nearby distractors (red outlines). The robot initiates motion early by moving to a \emph{preview} pose and then \emph{commits} to the predicted object only when the hand enters the close commit region. Yellow box: grasping motion. Top: VR interaction. Bottom: real robot execution.}
    \label{fig:robot_movement}
\end{figure}

We use UR10e manipulator and Robotiq-85 gripper in our work. 
All preview motions move the robot TCP to a fixed hover height of $h_{\mathrm{clear}} = 0.3$\,m above the preview positions predicted in \cref{subsec:policy_logic}, while maintaining a vertical, downward-facing tool pose. The robot interpolates between successive waypoints using straight-line Cartesian motion. For grasp execution, it descends vertically to a grasp height computed from the object pose (via ArUco \cite{ArUco} tracking) and known object dimensions. It then closes the gripper, and retreats to $h_{\mathrm{clear}}$. Similarly, for the post-grasp phase, the robot descends to a release height computed from the slot pose and geometry, opens the gripper, and retreats. The sequence of robot motions for the pre-grasping phase is illustrated in \cref{fig:robot_movement}.

\section{Evaluation}
\label{sec:evaluation}
We evaluate AHEAD in two complementary settings: (i) offline intent prediction on the collected dataset, and (ii) an user study that measures the benefits of proactive assistance during teleoperation.

\subsection{Intent prediction evaluation and ablations}
\label{sec:eval_offline}





\begin{table}[b!]
\centering
\setlength{\tabcolsep}{1.6pt}
\renewcommand{\arraystretch}{1.00}
\caption{Offline intent evaluation across phases.
\textbf{s} denotes seconds and \textbf{m} denotes meters.}
\label{tab:policy_compare}
\scriptsize
\resizebox{\columnwidth}{!}{
\begin{tabular}{lccccc|ccccc}
\toprule
\textbf{Method} &
\multicolumn{5}{c|}{\textbf{Obj}} &
\multicolumn{5}{c}{\textbf{Slot}} \\
\cmidrule(lr){2-6}\cmidrule(lr){7-11}
&
\makecell[c]{Acc$\uparrow$\\(T1/T3)} &
\makecell[c]{T1 Cross$\uparrow$\\(s/m)} &
\makecell[c]{T3 Cross$\uparrow$\\(s/m)} &
\makecell[c]{Lead$_{50}\uparrow$\\(s/m)} &
\makecell[c]{Flip$\downarrow$\\(/s)} &
\makecell[c]{Acc$\uparrow$\\(T1/T3)} &
\makecell[c]{T1 Cross$\uparrow$\\(s/m)} &
\makecell[c]{T3 Cross$\uparrow$\\(s/m)} &
\makecell[c]{Lead$_{50}\uparrow$\\(s/m)} &
\makecell[c]{Flip$\downarrow$\\(/s)} \\
\midrule
Nearest &
0.52/0.70 & 0.28/0.15 & 0.49/0.26 & 0.68/0.33 & \textbf{0.00} &
0.63/0.74 & 0.92/0.24 & 1.12/0.40 & 1.12/0.35 & \textbf{0.01} \\
\midrule
Unified &
0.75/0.88 & \textbf{0.60/0.31} & 0.98/0.54 & 0.86/0.48 & 0.06 &
0.75/0.90 & 1.03/0.27 & 1.58/0.65 & 1.35/0.52 & 0.08 \\
Decoupled &
\textbf{0.76/0.90} & 0.59/0.30 & \textbf{1.04/0.62} & \textbf{0.90/0.51} & 0.06 &
\textbf{0.76/0.91} & \textbf{1.05/0.30} & \textbf{1.62/0.68} & \textbf{1.38/0.54} & 0.09 \\
\bottomrule
\end{tabular}
}
\end{table}

We evaluate offline intent prediction using a \emph{5-fold leave-one-subject-out} (LOSO) protocol: in each fold, we train on data from four participants and test on the held-out participant (200 sequences per participant, 1{,}000 sequences total). We generate evaluation samples by extracting sliding windows of length $T$ with stride 5, using the same interpolation and 0.5 s windowing scheme as in training (see \cref{sec:problem_setup}). We compare AHEAD to the \textit{Nearest} heuristic, which predicts the intended object/slot as the candidate whose centroid is closest to the current palm position. In our offline evaluation, each scene contains $7.43 \pm 1.65$ objects and $7.30 \pm 1.67$ slots (mean $\pm$ SD), generated as described in \cref{subsec:scene_generation}.

\noindent\textbf{Ablation settings:}
\textbf{Unified vs.\ decoupled phase modeling.} We compare our final \textbf{decoupled} design (\cref{sec:method_model}), against a \textbf{unified} design. For the unified variant, we leverage multi-task learning and train a single model across both phases: the hand encoder is jointly optimized with both the object and slot branches, and supervision is routed by phase (PreGrasp windows train the object branch and CrossOver/PostGrasp windows train the slot branch). Instead, our decoupled variant trains two separate, phase-specific models: a pre-grasp object model and a post-grasp slot model. 

\noindent\textbf{Metrics:}
\textbf{Acc (Top 1 \& Top 3)} reports Top1 and neighbor-aware Top3 accuracy, with Top3 counting a prediction as correct if it falls within the ground-truth target's 0.3m neighborhood. These metrics measure model correctness.
\textbf{T1 Cross} and \textbf{T3 Cross} report the earliest time-to-event and distance-to-target at which the smoothed Top1 and Top3 accuracy curves exceed 0.8, thereby measuring how early predictions become reliable. \textbf{Lead$_{50}$ (s/m)} measures how early the model first becomes \emph{stably} correct: we report the median time and distance at the start of the first streak of six consecutive correct windows (0.1s at 60Hz). \textbf{Flip (/s)} is the rate of correct$\leftrightarrow$incorrect switches after this point, with lower values indicating greater stability. Ideally, a model should have a large Lead$_{50}$ and low flip (early and stable correctness).

\noindent\textbf{Results and model choice:}
Across both phases, learned predictors outperform \textit{Nearest} in correctness and reliability (see \cref{tab:policy_compare}). The Top1/3 drops for the \textit{Nearest} are smaller in the slot phase than the object phase because slots are more spatially separated from each other than objects, reducing nearest-neighbor ambiguity during placement. Relative to \textit{Decoupled}, \textit{Nearest} suffers large \emph{relative} accuracy drops: Obj Top1/Top3 $=-31.6\%/-22.2\%$, Slot Top1/Top3 $=-17.1\%/-18.7\%$. While it can appear stable by tracking instantaneous proximity, it is not anticipatory and becomes correct only near the target, yielding weaker T1/T3 Cross and Lead$_{50}$. Decoupled crosses Top-3 at larger distances by 0.36m (Obj: 0.62 vs.\ 0.26) and 0.28m (Slot: 0.68 vs.\ 0.40), enabling earlier proactive motion.



\noindent\textbf{Object phase:}
We select \textit{Decoupled} over \textit{Unified} because it improves the correctness--anticipation trade-off (higher Acc and Lead) without reducing temporal stability (similar Flip). Moreover, Decoupled reaches the Top3 reliability threshold earlier and at a greater hand-to-target distance, resulting in shorter robot reaction times.

\noindent\textbf{Slot phase:}
We observe the same pattern. \textit{Decoupled} improves accuracy and achieves earlier Lead than \textit{Unified}, with only a marginal increase in Flip. Overall, Decoupled provides more reliable proactive triggering with essentially unchanged stability, so we adopt it for both phases.

\noindent\textbf{Reliability Analysis of T1/T3 Cross:}
We report \emph{crossing-bin accuracy}, i.e., the observed accuracy in the first bin whose empirical accuracy reaches the 0.8 threshold used to define T1/T3 Cross. For each such bin, we denote the observed fraction of correct predictions by $\hat p$. Bracketed values denote Wilson 95\% confidence intervals (CIs) ~\cite{wilson}, which quantify uncertainty for the underlying binomial accuracy and remain well behaved for finite sample sizes, including proportions near 0 or 1. At their respective crossing bins (distance-based), the \textbf{Decoupled} object predictor attains Top1 $\hat p=0.815$ $[0.787, 0.840]$ and Top3 $\hat p=0.813$ $[0.773, 0.847]$. The slot predictor attains T1 Cross $\hat p=0.824$ $[0.796, 0.849]$ and T3 Cross $\hat p=0.828$ $[0.798, 0.855]$. These crossing-bin estimates show that the selected policy thresholds in \cref{sec:method_policy} place the controller at the intended reliability level suitable for proactive control. The remaining uncertainty (Wilson CIs $\leq 0.8$) motivates conservative distance-gated preview rather than immediate execution.

\noindent\textbf{Discussion:} 
These results suggest that a single shared hand encoder does not yield consistent gains across phases, likely due to phase-dependent motion characteristics: pre-grasp reaching often follows a relatively consistent approach (typically forward and downward toward the object), whereas post-grasp transport and placement are more variable and can include lateral and upward motions with corrective adjustments toward the slot. Separate encoders can therefore specialize to these phase-specific cues.
We assess cross-user generalization with LOSO cross-validation. The model achieves high accuracy across folds, and the reliability-threshold analysis remains satisfied on held-out users, indicating that AHEAD’s intent predictions generalize across users. This aligns with the user study in \cref{sec:eval_userstudy}, where we successfully deploy the model in teleoperation with unseen users and a different scene, whilst keeping $d^{\mathrm{Top1}}_{\mathrm{acc}=0.8}$ and $d^{\mathrm{Top3}}_{\mathrm{acc}=0.8}$ unchanged. Nevertheless, the dataset remains small and motivates larger-scale data collection to better characterize generalization across users and scenes.

\subsection{Controlled VR User Study Setup}
\label{sec:eval_userstudy}

\begin{figure}[h!]
    \vspace{4pt}
    \centering
    \includegraphics[width=\columnwidth]{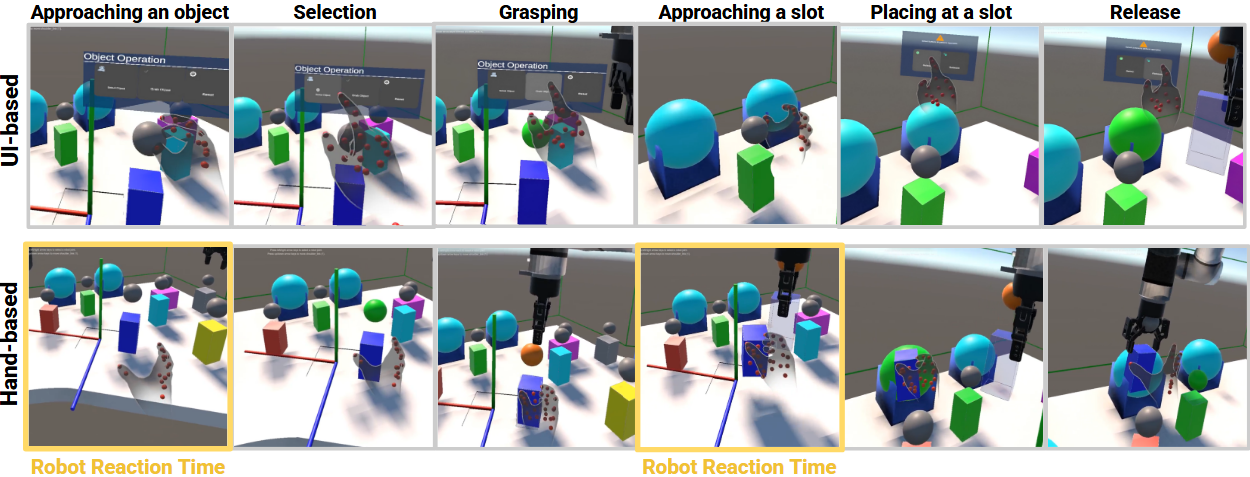}
    \caption{Overview of four interaction methods. UI requires explicit object and slot selection, while Dwell, Nearest, and AHEAD let the operator command the robot through natural hand motion (approach, grasp, transport, release). Dwell, Nearest, and AHEAD share a similar interaction flow and differ only in the object selection mechanism.}
    \label{fig:baseline_explanation}
    
\end{figure}

\noindent\textbf{User study design:}
We conducted a within-subject study where each participant completed the same pick-and-place task under four teleoperation-interface conditions, with condition order counterbalanced via a Latin-square to mitigate learning and fatigue effects. For every condition, we instructed participants to move six objects into six pre-assigned target slots (object $i \rightarrow$ slot $i$), with object and slot locations fixed across conditions. We used the same robot motion primitives across all conditions on the real robot (UR10e), with AHEAD applying a condition-specific adjustment to enable preview motion (see \cref{subsec:motion_primitives}). Importantly, the user-study scene layout differs from the data-collection setup, so the intent model was not overfitted to the study setup. Visual cues show interaction state: selected object indicators/slots turn green and grasped objects turn orange.
\cref{fig:baseline_explanation} shows the four teleoperation conditions used in our comparison. Dwell, Nearest, and AHEAD are hand-based methods.
\begin{compactitem}
\item \textbf{UI-based:} Object selection, grasp execution, slot selection, and release execution are all triggered via UI panels. Operators open menus by pinching near each object's sphere and inside a slot region (top row of \cref{fig:baseline_explanation}).
\item \textbf{Hand-based (3 variants):} We evaluate three hand-driven baselines that differ in how object/slot targets are selected. \textit{Dwell} selects an object after a 1s dwell near it and selects a slot when the grasped object enters a slot region, while grasp and release follow natural hand motions. \textit{Nearest-object} reactively selects, at each frame, the object closest to the user’s palm, while using the same slot-selection, grasp, and release interactions as \textit{Dwell}. \textit{AHEAD} follows the same interaction flow as \textit{Nearest-object}, but predicts object and slot intent from hand motion and scene context, allowing the robot to initiate motion earlier.
\end{compactitem}

\noindent We conduct a post-condition survey with 4--7 questions per metric (Naturalness, Efficiency, Understanding, and Trust), in addition to NASA-TLX~\cite{NASATLX}. Naturalness measures how natural robot control feels. Efficiency measures how quickly and smoothly participants complete the task with minimal extra steps or corrections. Understanding measures how well participants can interpret the robot’s current state and motion. Trust measures how willing participants are to rely on the robot’s future behavior without constant monitoring. \emph{React} in \cref{tab:user_survey} measures the time from the operator’s hand leaving a fixed start point (palm $>0.1$\,m from the start position) to the robot’s first motion toward the correct target, ensuring consistent measurement across participants and conditions.

\noindent\textbf{Participants}
We recruit $N{=}15$ participants (5 females, 10 males; age: $25.0\pm4.5$ years). Self-reported familiarity scores (0--10 Likert scale) are $6.87\pm2.88$ for VR, $5.60\pm2.85$ for robotics, and $4.93\pm3.27$ for teleoperation.

\begin{table}[t!]
\vspace{6pt}
\centering
\setlength{\tabcolsep}{2.1pt}
\renewcommand{\arraystretch}{1.05}
\caption{User-study ratings by condition (mean$\pm$SD) with omnibus within-subject effects. B1--B4 denote UI, Dwell, Nearest, and AHEAD. \emph{React} denotes robot reaction time.}
\label{tab:user_survey}
\scriptsize
\begin{tabular}{lcccc@{\hspace{2pt}}!{\vrule width 0.6pt}@{\hspace{2pt}}cccc}
\toprule
& \multicolumn{4}{c@{\hspace{2pt}}!{\vrule width 0.6pt}@{\hspace{2pt}}}{\textbf{Teleop. Interface Conditions}} & \multicolumn{4}{c}{\textbf{Friedman}} \\
\cmidrule(lr){2-5}\cmidrule(lr){6-9}
\textbf{Cat.} & \textbf{B1} & \textbf{B2} & \textbf{B3} & \textbf{B4} &
$\boldsymbol{\chi^2}$ & $\mathbf{p}$ & $\mathbf{W}$ & \textbf{Sig.} \\
\midrule
Naturalness   & 4.2$\pm$2.5 & 5.1$\pm$1.7 & 7.8$\pm$2.0 & \textbf{8.6}$\pm$0.8 & 29.84 & $1.0\mathrm{e}{-6}$ & 0.66 & \checkmark \\
Efficiency    & 4.7$\pm$2.7 & 5.2$\pm$1.6 & 7.7$\pm$1.6 & \textbf{8.1}$\pm$1.5 & 17.78 & $4.89\mathrm{e}{-4}$ & 0.40 & \checkmark \\
Understanding & 7.8$\pm$1.7 & 7.0$\pm$2.1 & 8.1$\pm$1.4 & \textbf{8.4}$\pm$0.9 &  6.96 & 0.07327 & 0.16 & -- \\
Trust         & 6.5$\pm$1.3 & 5.9$\pm$2.1 & 7.7$\pm$1.4 & \textbf{7.9}$\pm$1.5 & 16.55 & $8.74\mathrm{e}{-4}$ & 0.37 & \checkmark \\

\specialrule{0.6pt}{0.6ex}{0.6ex}
$React_{obj}$[s]
              & 3.7$\pm$1.0 & 3.0$\pm$0.6 & 1.5$\pm$0.4 & \textbf{0.9}$\pm$0.3
              & 40.04 & $1.04\mathrm{e}{-8}$ & 0.89 & \checkmark \\
$React_{slot}$[s]
              & 4.5$\pm$0.9 & 2.3$\pm$0.9 & 2.2$\pm$0.6 & \textbf{0.8}$\pm$0.5
              & 41.00 & $6.54\mathrm{e}{-9}$ & 0.91 & \checkmark \\
\bottomrule
\end{tabular}
\end{table}

\subsection{User Study Results}

We report user survey results and robot reaction time in \cref{tab:user_survey}. We analyzed within-subject Likert ratings (0-10) using Friedman tests ($N{=}15$ participants, $k{=}4$ conditions, $\chi^2(3)$, $p$), where the degrees of freedom are $k{-}1{=}3$, and report Kendall's $W$ as an effect size \cite{kendall}. For significant omnibus effects, we ran Wilcoxon signed-rank post-hoc tests with Holm correction \cite{holm}. Naturalness differed strongly across baselines ($\chi^2(3){=}29.84$, $p{=}1.0\mathrm{e}{-6}$, $W{=}0.66$): \textit{AHEAD} and \textit{Nearest} were rated significantly more natural than \textit{UI} and \textit{Dwell} (all $p_{\mathrm{Holm}}{<}.05$), with no significant difference between \textit{AHEAD} and \textit{Nearest}. Efficiency also differed ($\chi^2(3){=}17.78$, $p{=}4.89\mathrm{e}{-4}$, $W{=}0.40$), again favoring \textit{AHEAD} and \textit{Nearest} over \textit{UI} and \textit{Dwell} (all $p_{\mathrm{Holm}}{<}.05$), with no difference between \textit{AHEAD} and \textit{Nearest}. Trust showed an omnibus effect ($\chi^2(3){=}16.55$, $p{=}8.74\mathrm{e}{-4}$, $W{=}0.37$). The only pairwise contrast surviving Holm correction was \textit{Nearest} $>$ \textit{Dwell} ($p_{\mathrm{Holm}}{=}0.017$), while other comparisons did not. Understanding did not differ across baselines ($\chi^2(3){=}6.96$, $p{=}0.073$, $W{=}0.16$). 
Finally, the participants completed 100\% of trials for all 4 conditions, so task completion does not distinguish the conditions. The key difference is therefore reaction time, where AHEAD achieves the shortest among the four. Although AHEAD did make occasional intent prediction errors, these did not prevent task completion or negate its reaction-time advantage over the other methods. This suggests that its predictions are reliable enough in practice to support earlier and more consistent robot motion.

\noindent\textbf{NASA-TLX:} We analyzed within-subject NASA-TLX ratings (0--100) using Friedman tests and report Kendall’s $W$. Mental demand differed across baselines (B1--B4: 48.5$\pm$21.4, 46.1$\pm$20.9, 35.6$\pm$26.6, 31.7$\pm$29.2. $\chi^2(3){=}11.19$, $p{=}0.0107$, $W{=}0.27$) with lower demands for \textit{Nearest}/\textit{AHEAD} than \textit{UI}/\textit{Dwell}. Physical demand showed a similar effect (50.2$\pm$21.1, 51.6$\pm$21.8, 39.0$\pm$26.1, 31.3$\pm$31.0. $\chi^2(3){=}14.07$, $p{=}0.00281$, $W{=}0.33$). Effort also differed (44.7$\pm$24.4, 54.1$\pm$21.2, 37.9$\pm$27.5, 31.0$\pm$24.9. $\chi^2(3){=}9.79$, $p{=}0.0204$, $W{=}0.23$) and post-hoc tests identified lower effort for \textit{AHEAD} than \textit{Dwell} ($p_{\mathrm{Holm}}{=}0.0345$). Self-rated performance was high across conditions (84.3$\pm$15.0, 83.3$\pm$16.0, 90.0$\pm$9.0, 92.0$\pm$7.7. $\chi^2(3){=}7.00$, $p{=}0.0719$, $W{=}0.17$) since the task was simple and the study focused on comparing interaction methods rather than task success. Overall, these results indicate that \textit{AHEAD} improves the teleoperation experience by making interaction feel more natural and efficient, thus reducing perceived workload without reducing users’ understanding of the robot’s behavior. Compared to Nearest, AHEAD preserves similarly strong subjective ratings while enabling anticipatory motion initiation.

\section{Conclusions and Future Work}
\label{sec:conclusion}
We present \textbf{AHEAD}, an anticipatory VR teleoperation system that predicts pick-and-place intent from a short history of 3D hand, head pose, and a scene context. Across offline evaluation and a user study, AHEAD reduced robot reaction time through proactive motion initiation while maintaining low operator workload and stable behavior under dynamic human motion.
Our current system has several limitations. First, it currently relies on ArUco-based tracking, which simplifies object pose estimation but restricts real-world deployment. Second, our dataset remains limited in scale, which constrains how fully we can assess cross-user and cross-layout generalization. Finally, AHEAD predicts target identity but not approach geometry, which limits orientation-aware grasping (e.g., selecting a cube face or grasp direction). Future work will address these limitations by replacing marker-based tracking with markerless vision-based pose estimation, scaling data collection, and extending prediction to approach direction.

\section{Acknowledgement}
\label{sec:acknowledge}

This material is based upon work supported by the National Science Foundation (NSF) under Awards 2514364 and 2528653 and by the Industrial Technology Innovation Program (P0028404) funded by the Ministry of Trade, Industry and Energy of the Republic of Korea. The views, opinions and conclusions expressed in this work are those of the authors and do not reflect those of the funding agencies.

\bibliographystyle{IEEEtran}
\bibliography{IEEEabrv,refs}

\end{document}